\documentclass[journal]{IEEEtran}
\usepackage{amsmath,amsfonts}
\usepackage{algorithmic}
\usepackage{algorithm}
\usepackage{graphicx}      % include this line if your document contains figures\usepackage{array}
\usepackage[caption=false,font=normalsize,labelfont=sf,textfont=sf]{subfig}
\usepackage{textcomp}
\usepackage{stfloats}
\usepackage{url}
\usepackage{verbatim}
\usepackage{graphicx}
\usepackage{cite}
\newcommand{\RNum}[1]{\uppercase\expandafter{\romannumeral #1\relax}}
\usepackage[inline]{enumitem}
\usepackage[dvipsnames]{xcolor}
\newcommand{\Tulga}[1]{\textcolor{Red}{[Tulga: #1]}}
\newcommand{\Siyuan}[1]{\textcolor{Green}{[Siyuan: #1]}}
\newcommand{\Congkai}[1]{\textcolor{Cyan}{[Congkai: #1]}}
\newcommand{\Discussion}[1]{\textcolor{Blue}{[Discussion: #1]}}

\usepackage[colorlinks=true, allcolors=blue]{hyperref}

\usepackage[dvipsnames]{xcolor}

\usepackage{ifthen}
\usepackage[normalem]{ulem} % for spacing issue for "subequations"
\usepackage{etoolbox}% for spacing issue for "subequations"

\newboolean{showmodification}
\setboolean{showmodification}{true}
\makeatletter
\ifthenelse{\boolean{showmodification}}{
}
{
 	\AtBeginDocument{\let\hl\@firstofone}
	\AtBeginDocument{\let\hlmath\@firstofone}	
	\renewcommand{\Tulga}[1]{}
	\renewcommand{\Siyuan}[1]{}
	\renewcommand{\Congkai}[1]{}
    \renewcommand{\Discussion}[1]{}
}
\makeatother

\makeatother

\begin{document}

\title{Cyber Racing Coach: A Haptic Shared Control Framework for Teaching Advanced Driving Skills}

\author{Congkai Shen$^\dagger$, Siyuan Yu$^\dagger$, Yifan Weng, Haoran Ma, Chen Li, Hiroshi Yasuda, James Dallas, Michael Thompson, John Subosits, Tulga Ersal$^*$
\thanks{\textcolor{red}{This work has been submitted to the IEEE for possible publication. 
Copyright may be transferred without notice, after which this version may no longer be accessible.}}
\thanks{$^\dagger$These authors contributed equally to this work.}
\thanks{* Corresponding author}
\thanks{C. Shen, S. Yu, Y. Weng, H. Ma, C. Li and T. Ersal are with the Department of Mechanical Engineering, University of Michigan, Ann Arbor, MI 48109. (email: \{cosh, johnysy, wenyifan, hrma, clicli, tersal\}@umich.edu)}
\thanks{H. Yasuda, J. Dallas, M. Thompson, J. Subosits are with Toyota Research Institute, Los Altos, CA 94022. (email: \{hiroshi.yasuda, james.dallas, michael.thompson, john.subosits\}@tri.global)}
}

% \markboth{IEEE Transactions on Systems, Man, and Cybernetics: Systems,~Vol.~?, No.~?, ?~2024}{}


\maketitle

\begin{abstract}
This study introduces a haptic shared control framework designed to teach human drivers advanced driving skills.	
In this context, shared control refers to a driving mode where the human driver collaborates with an autonomous driving system to control the steering of a vehicle simultaneously. 
Advanced driving skills are those necessary to safely push the vehicle to its handling limits in high-performance driving such as racing and emergency obstacle avoidance.	
Previous research has demonstrated the performance and safety benefits of shared control schemes using both subjective and objective evaluations.
However, these schemes have not been assessed for their impact on skill acquisition on complex and demanding tasks. 
Prior research on long-term skill acquisition either applies haptic shared control to simple tasks or employs other feedback methods like visual and auditory aids.
To bridge this gap, this study creates a cyber racing coach framework based on the haptic shared control paradigm and evaluates its performance in helping human drivers acquire high-performance driving skills.
The framework introduces (1) an autonomous driving system that is capable of cooperating with humans in a highly performant driving scenario; 
and (2) a haptic shared control mechanism along with a fading scheme to gradually reduce the steering assistance from autonomy based on the human driver's performance during training.
Two benchmarks are considered: self-learning (no assistance) and full assistance during training.
Results from a human subject study indicate that the proposed framework helps human drivers develop superior racing skills compared to the benchmarks, resulting in better performance and consistency.
% , even when starting from the same skill level.

\end{abstract}

\begin{IEEEkeywords}
Shared Control, model predictive control, human subject test
\end{IEEEkeywords}

\section{Introduction}\label{sec:intro}

Advanced driving skills refer to a set of competencies that go beyond basic driving abilities in terms of situational awareness, hazard perception, risk management, and vehicle handling \cite{isler2011effects}. 
They are crucial in high-performance driving tasks such as racing, and can also improve safety in everyday driving \cite{washington2011european,isler2011effects}.
However, they can be challenging to learn due to several reasons:	
(1) \textbf{Complexity:} Advanced driving skills require proficiency in vehicle control, hazard perception, and risk management, making them inherently complex and difficult to master quickly.
For instance, accurately gauging handling limits and responding appropriately demands extensive practice under conditions not experienced regularly in everyday driving.	
(2) \textbf{Risk:} Advanced driving involves higher risk levels than basic driving. 
This can intimidate learners and requires time to build necessary confidence and decision-making skills.	
(3) \textbf{Cognative \& physical Demanding:} Mastering these skills requires high mental workload, dexterity, and coordination, which can be challenging for learners.

On the other hand, autonomy algorithms have been witnessed to safely explore high-performance driving limits \cite{wurts2021collision, betz2023tum, goh2016simultaneous, wischnewski2022indy, dallas2021terrain}. 
Therefore, it is possible to conceive a framework that leverages highly performant autonomy algorithms to teach human drivers advanced driving skills that enable precise handling at the dynamic limits of vehicles.
However, as the literature review below shows, such a framework has not yet been created. 
This study addresses this gap by using racing as a context for assessing human drivers' advanced driving skills.

\subsection{Background}

% Literature Review on using all sorts of feedback (audio, visual) for advanced driving methods
    % ~150 words
% Over the past few decades, expansions occurred in the types and usage of vehicle simulators \cite{DrivSimRev}. Simulators for car, truck, ship, and submarine emerged in addition to flight simulators. Cost and risk are also no longer the only concern behind simulators, as entertainment and scientific use are added to the usage of vehicle simulators \cite{DrivSimRev}. In recent years, simulators for racing applications were widely applied in the motorsports industry for professional drivers to train and prepare for the track \cite{RaceSimPrep}. Advantages of training with simulators include the ability to replicate real world driving condition, reduce training cost, and improve driving skills \cite{RaceSimPrep,F2Sim}.

% Despite the extensive use of racing simulators to train drivers and the various driving aids such as the racing line and stability control embedded in the systems, these simulators does not provide an active learning experience to the user \cite{RaceSimTrain}. Until now, few studies were conducted to examine the effectiveness of integrating feedback models to simulators. Bugeja et al. developed a telemetry-based feedback model using measurement of various vehicle states to provide real-time visual and audio suggestions to the driver, which showed its effectiveness by a better performance of the testing group with respect to the control group \cite{RaceSimTrain}.

In recent advancements in racing training programs and games, it has been established that providing visual and auditory feedback is beneficial for novice drivers \cite{tonnis2007visual,yang2017effect}.
In \cite{Keith2017Telemetry}, visual feedback, such as illustrating the racing line on the track, assists drivers in determining the optimal path to steer around corners. 
% Aside from common auditory instructions, researchers in \cite{brunt2017effects} devised a mechanism that emits sounds corresponding to the degree of deviation of the vehicle from the ideal slip angle, providing a steering cue.
In \cite{scacchi2021interactive}, the integration of auditory and visual feedback components culminates in the creation of an immersive learning environment tailored for novice players. 
Nevertheless, these feedback modalities are often advisory, meaning that they only provide suggestions for humans instead of directly interacting with their behavior. 
With a motivation to provide physical interactions with humans, the present study examines the viability of haptic feedback in instructing advanced driving techniques.

Haptic feedback allows for physical interactions between the human and autonomous driving system, enabling the human to feel the intentions of autonomy in real time.
This continuous mode of collaboration enhances seamless transitions of control authority, which is particularly crucial for applications such as shared control in semi-autonomous driving \cite{zilberstein2015building}. 
Here, haptic shared control means that both the human driver and autonomy can exert a torque on the steering wheel at the same time, reflecting their respective intentions to steer. 
This shared control mechanism allows both agents to communicate their control intentions through the steering wheel.
This mutual feedback forms a closed-loop system, enhancing coordination and responsiveness, and turning the steering wheel into a negotiation interface.
In \cite{griffiths2005sharing}, haptic shared control is shown to reduce the visual demand and shorten the reaction time while performing better in the driving test. 
These benefits make haptic shared control a valuable tool in enhancing driver safety and efficiency.
It has also been shown to reduce the workload of the human driver in performing ordinary driving tasks such as lane keeping  \cite{tada2016simultaneous,brandt2007combining,weng2020design, wang2017effect, luo2021workload}. 

Prior work has primarily focused on characterizing the immediate benefits of haptic shared control. 
Most studies have concentrated on immediate improvements in task performance and safety, without exploring the longer term implications and potential for driver training.
In other words, haptic shared control has been initially used as a driver assistance tool rather than as a teaching aid.

While haptic shared control effectively supports humans in real-time, there is considerable potential for its further development as a training mechanism for advanced driving skills. 
The rationale for this hypothesis is grounded in the applications of haptic shared control in normal driving tasks \cite{lee2014combining, marchal2010effect, petermeijer2015should, lee2010effects,de2011preparing,de2011effect}. 
In \cite{marchal2010effect, de2011preparing}, continuous haptic feedback was provided to guide drivers in a steering task, helping them to follow lanes. 
Haptic feedback has also been utilized as a continuous disturbance to help drivers improve their skills \cite{lee2014combining, lee2010effects}.
Notably, in \cite{lee2014combining}, a fading scheme is created to gradually fade away the torque assist to help the human's learning process.
Comparatively, as noted in \cite{de2023shared}, consistent assistance without a fading mechanism may lead to deskilling, as automation can hinder long-term skill retention rather than enhance it.
Haptic feedback can also be provided in a bandwidth-limited manner, activating only when certain system states exceed a threshold.
In \cite{petermeijer2015should, de2011effect}, researchers demonstrate that both continuous guidance and bandwidth feedback can effectively enhance driver performance. 
Although haptic feedback has demonstrated success in teaching longer-term skills to human drivers in driving scenarios, it has only been applied in relatively simple contexts. 
The potential for utilizing haptic shared control in more complex and performance-oriented driving scenarios has not yet been explored.

To effectively teach human drivers advanced driving skills via haptic shared control, it is essential to have a highly performant autonomous system that not only possesses such skills itself but also can seamlessly collaborate with humans.	
In \cite{anderson2010optimal}, an optimal control-based algorithm is used to track the path with its own steering commands, and the final steering command is a weighted sum of both human and autonomy inputs.
In \cite{benloucif2017new, benloucif2019cooperative}, researchers hired a human expert that is utilized as a source of autonomy that generates haptic feedback. 
To avoid the inconsistency of human experts, other researchers introduced a 
% Legendre-Gaussian-Radau based \Tulga{unnecessary detail at this point}
model predictive control (MPC) formulation within the haptic shared control framework \cite{weng2020design}.
Although such studies have shown that autonomy and human drivers can collaborate, the autonomous systems used in current haptic shared control frameworks are not performant enough for advanced driving scenarios such as racing.

Recent efforts resulted in advanced autonomous systems capable of achieving performance comparable to the best human drivers.
Specifically in the context of racing, machine learning algorithms, such as reinforcement learning, have been widely used due to their ability to find optimal policies and their general applicability \cite{balaji2019deepracer, song2023reaching, remonda2021formula}. 
% However, machine learning algorithms are often trained without human intervention, and the addition of human disturbances can result in sub-optimal behavior.	
Other works  utilize feedforward and feedback controllers to calculate the steering, throttle and braking commands over the racetrack \cite{subosits2019racetrack, talvala2011pushing, kapania2016sequential}. 
MPC has also been popular because it captures the vehicle model and maps control actions based on this knowledge \cite{dallas2023hierarchical, betz2022autonomous, spielberg2023learning, hewing2020learning, williams2018information}. 
In \cite{talbot2023shared},  MPC is used in a shared control mode, where it manages the throttle and braking positions while the human driver controls the steering.
Nevertheless, these solutions either rely on a pre-defined global trajectory to inform the local controller or use learned objectives for control.	
With human intervention, the vehicle can easily deviate from the original lane in trajectory tracking methods or from the safe states set in learning-based MPC, which can cause infeasible solutions. 

In \cite{yu2025spatial}, a safe envelope-based MPC method is introduced, which neither requires training nor relies on pre-defined trajectories when planning dynamically feasible paths on the track. However, it is designed to operate autonomously, without consideration for shared control with human intervention.

\subsection{Original Contributions}

% Although previous research has demonstrated significant success in utilizing haptic shared control to assist humans in completing driving tasks, the state-of-the-art methods have not yet proven effective in challenging scenarios that require specific expertise, especially when considering long-term benefits. 
In light of the background review above, the potential benefits of haptic shared control in helping human drivers acquire advanced driving skills is identified as an unexplored but important domain.
To explore this potential, this paper presents and evaluates a novel haptic shared control framework aimed at teaching humans advanced driving skills. 
% This framework is designed to address the identified gap by offering haptic feedback, thereby enhancing the effectiveness of long-term training and helping learners develop the necessary skills.
% 
The original contributions of this work are: 
\begin{enumerate}
    \item A real-time Model Predictive Control (MPC) framework capable of optimizing vehicle trajectories online without requiring a predefined path while being robust to human intervention.	
    \item A haptic shared control framework using a fading scheme to smoothly transition control authority by fading away the autonomy assistance based on human driver's performance during training.
    % \item A haptic shared control framework that can transition control authority smoothly between human drivers and autonomy. \Tulga{What are we claiming as contribution here? As written, and given the literature we reviewed in the background, it looks like there is no contribution in this statement.}
    % \item Performance metrics used to evaluate advanced driving skills of human subjects. \Tulga{This is part of our methodology, but does not qualify as a contribution.}
    % \item A fading scheme that fades away the assistance from autonomy based on the human driver's performance during training.
    \item Validation of the efficacy of the proposed framework for teaching advanced driving skills through a human subject study, with self-learning and full assistance teaching methods as the benchmarks.
\end{enumerate}

The remainder of the paper is organized as follows. 
Sec. \ref{sec:methods} covers the methodology, including the design of autonomy, the haptic shared control framework, performance evaluation metrics, and the fading scheme in training. 
Sec. \ref{sec:exp_setup} describes the hardware and software setup for the experiment, the test procedure, and the demographics of the human subjects. 
Sec. \ref{sec:exp_res} discusses the results of the human subject study and further highlights the benefit of utilizing the proposed methods. 
Finally, Sec. \ref{sec:conclusion} concludes the study, summarizing the key findings and future directions of research.

\section{Methods}\label{sec:methods}
\begin{figure*}
\centering
\includegraphics[width=1.0\textwidth]{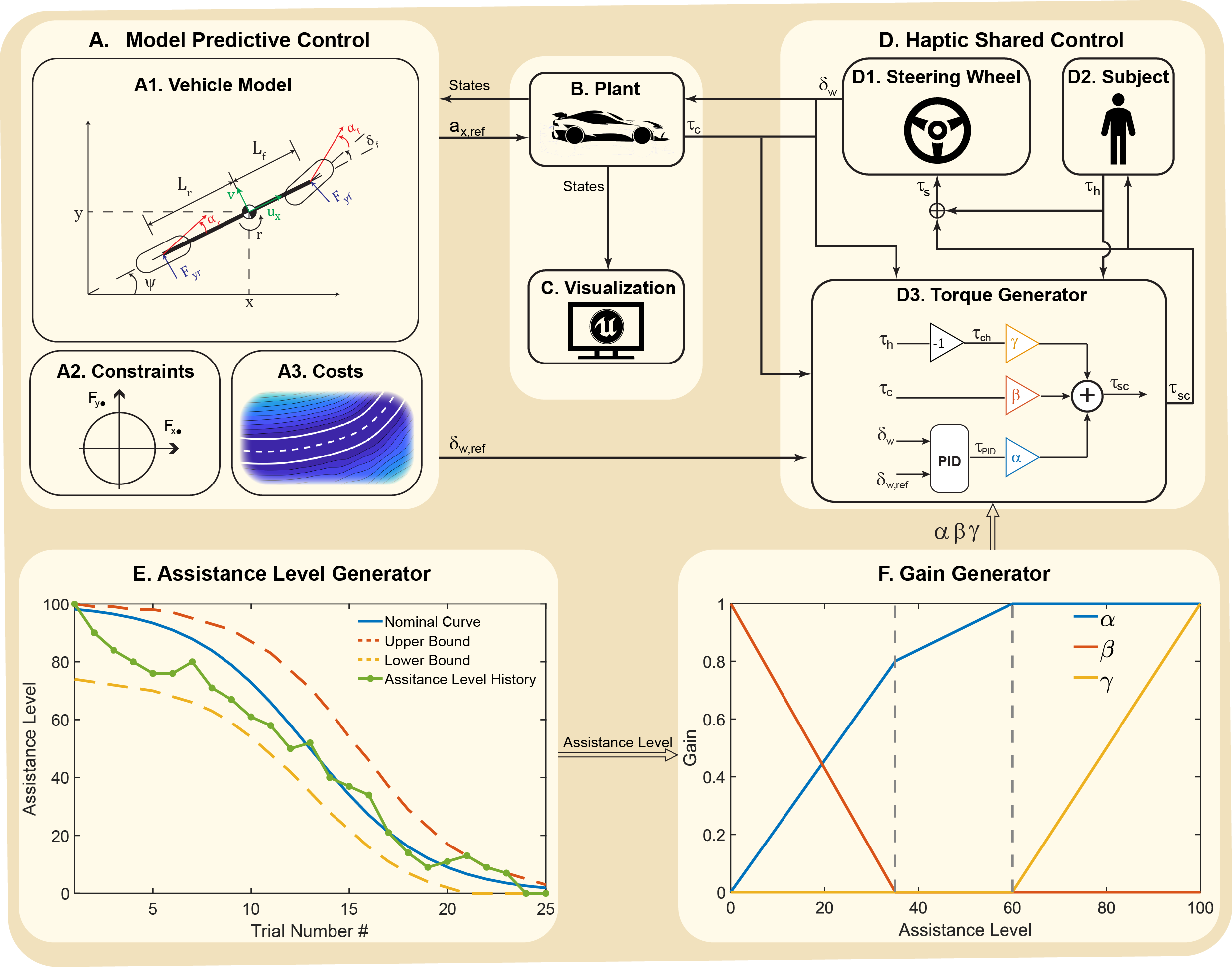}
\caption{The framework of shared control. Solid arrow represents real-time information transfer, while hollow arrow indicates information transfer only at the beginning of each run. (\textbf{A1}) 3 Degree of Freedom (DoF) bicycle model (see Sec.~\ref{sec:VD}). (\textbf{A2}) Hard constraint of friction circle limit. (\textbf{A3}) Surface plot of the potential field for the cost of boundary violation (see Sec.~\ref{sec:SC}). (\textbf{B}) Plant model for simulation. (\textbf{C}) Visualization in Unreal Engine 4 (see Sec.~\ref{sec:exp_setup}). (\textbf{D1}) Logitech G29 steering wheel with torque sensor (see Sec.~\ref{sec:exp_setup}). (\textbf{D2}) Human subject (see Sec.~\ref{sec:human_subject}). (\textbf{D3}) Schematic block diagram of steering torque generator. (\textbf{E}) Evolution of assistance level along the trial number of one subject guided by fading scheme (see Sec.~\ref{sec:FS}). (\textbf{F}) Design of the three gains $\alpha, \beta, \gamma$ as functions of the assistance level}
\label{fig:framework}
\end{figure*}

\subsection{Overall Framework} \label{sec:overall_framework}

Fig.~\ref{fig:framework} provides an overview of the proposed haptic shared control framework.
Starting with the block A, the vehicle states are fed as the initial condition to the autonomy, which optimizes the trajectories as detailed in Sec.~\ref{sec:MPC}. 
The resulting optimal acceleration $a_{x,\text{ref}}$ and  steering wheel angle $\delta_{w,\text{ref}}$ are provided as references to the blocks B and D3, respectively.
The human subject does not control the speed; therefore, $a_{x,\text{ref}}$ is sent directly to the plant. 
The haptic shared control block D outputs the actual steering wheel angle $\delta_w$ based on the optimal steering angle $\delta_{w,\text{ref}}$ and the centering torque $\tau_c$, as detailed in Sec.~\ref{sec:HSC}. 
At the start of each training trial, the assistance level generator (block E) determines the next assistance level for the human subject. 
The behavior of the assistance level generator varies depending on the training method. 
Fig.~\ref{fig:framework}-E illustrates the behavior of the fading-scheme training method, as detailed in Sec.~\ref{sec:FS}. 
Given the assistance level, the gains $\alpha$, $\beta$, and $\gamma$ in the torque generator (block D3) are adjusted accordingly, as depicted in Sec.~\ref{sec:HSC}. 
The plant and the visualization are explained in Sec.~\ref{sec:exp_setup}.

\subsection{Design of Autonomy}\label{sec:MPC}
The nonlinear Model Predictive Control (MPC) framework developed in \cite{yu2025spatial} is utilized as the autonomy for this study. 
It provides the steering and acceleration references used in the plant and torque generator.
This section summarizes the formulation for completeness, and the interested reader is referred to \cite{yu2025spatial} for further details and validation.

The following optimal control problem (OCP) is solved within the MPC over a receding horizon:
\begin{align}
\label{eq:cost} \mathop{\mathrm{minimize}}\limits_{\xi,\ \zeta}\quad \quad &J =\int_{t_0}^{t_f}\mathcal{I}[\xi(t),\zeta(t)]dt + \mathcal{G}[\xi(t_f)]\\
\label{eq:Dynamic}\text{subject to} \quad \quad &\dot{\xi}(t) = \mathcal{V}[\xi(t),\zeta(t)]\\
\label{eq:state}& \xi_\text{min}\leq \xi(t) \leq \xi_\text{max}\\
\label{eq:ctr}& \zeta_\text{min}\leq \zeta(t) \leq \zeta_\text{max}\\
\label{eq:tire_force_cons}& \mathbf{\mathcal{T_F}}[\xi]\leq \mathbf{0}
\end{align}
Here, $J$ is the cost function, $\mathcal{I}$ is the stage cost for each point along the predicted trajectory and $\mathcal{G}$ represents the cost-to-go.
The OCP is constrained to follow the vehicle dynamics $\mathcal{V}$, while the states $\xi$ and controls $\zeta$ are bounded as in Eq.~\eqref{eq:state} and ~\eqref{eq:ctr}. 
$\mathcal{T_F}$ represents the constraints for the tire force limits of the vehicle.

\subsubsection{Vehicle Dynamics}\label{sec:VD}

The vehicle dynamics model is developed and tested with respect to the real vehicle telemetry in \cite{yu2025spatial}. This model uses the traditional single-track approximation to combine the two tires on the front and rear axles as shown in Fig.~\ref{fig:framework}-A1. The model is tuned to represent the dynamics of the Toyota Research Institute LC 500 testing vehicle and it is capable of capturing the following characteristics of the vehicle:
\begin{enumerate*}
    \item rear wheel drive system;
    \item load transfer behavior;
    \item nonlinear tire forces;
    \item friction ellipse of tire forces.
\end{enumerate*}
This vehicle model is specially tailored for closed-loop optimization based control algorithms to balance  model fidelity and computational performance. See \cite{yu2025spatial} for further details.

\subsubsection{Hard Constraints} Hard constraints  imposed on the states and control inputs for vehicle safety and assumed actuator limits are similar to \cite{yu2025spatial}. Different from \cite{yu2025spatial}, a speed constraint is added to lower the task difficulty. Without this speed limit, even minor disagreements on the steering command between humans and autonomy would cause significant failures. This constraint is written as:
\begin{equation}
    \label{eq:ux_bound} u_{x_\mathrm{min}}\, \leq \, \, \,  u_x \leq \, u_{x_\mathrm{max}}
\end{equation}
where $u_{x_\mathrm{min}}=1$ m/s and $u_{x_\mathrm{max}}=27.5$ m/s. The minimum value is set to avoid the unstable response of the dynamical bicycle model at very low speeds.
Based on the pilot study, the maximum speed is set below the fully autonomous capability on the race track to provide a safety margin for human intervention. This limitation may affect the race line during high-speed turns but has minimal impact on hairpin turns.

Compared to \cite{yu2025spatial}, the hard constraints on envelopes are removed so that the algorithm is still capable of planning trajectories back to the track when the vehicle goes outside the boundaries due to human intervention.
% Otherwise, the violations of hard constraints would stop the solver from solving valuable solutions. 
% Please refer to \cite{yu2025spatial} for additional details on hard constraints.

\subsubsection{Soft Constraints}\label{sec:SC}: As in  \cite{yu2025spatial}, the cost function $J$ is expressed by four terms: 
\begin{equation}
    J = J_{\text{state}} + J_{\text{control}} + J_{\text{envelope}} + J_{\text{specific}}
\end{equation}

The soft constraint $J_{\text{envelope}}$ is defined to make the autonomy reliable by preventing infeasibility when it cooperates with humans.
\begin{equation}
    J_{\text{envelope}} = \sum_{i=1}^{n_\text{ds}} w_{\text{tube}}  \log\left(1 + e^{ - \theta_\text{hp}  (g_\text{sm} + g_{\text{envelope}}(\mathcal{T}(i)) } \right)
\end{equation}
Here $g_{\text{envelope}}(\mathcal{T}(i))$ is the value of envelope constraint described in \cite{yu2025spatial}, $g_\text{sm}$ is the safety margin, $\theta_\text{hp}$ is a hyper parameter and $w_{\text{tube}}$ is the weight. The soft constraint uses a soft plus function where the cost increases in a linear rate when the vehicle violates the bound and thus the gradient of this cost function can push the vehicle back to the circuit quickly.

The other modification from \cite{yu2025spatial} lies in the specific cost ($J_{\text{specific}}$). In this work, this cost is written as: 
\begin{equation}
    J_{\text{specific}} = J_{\text{go}} + J_{\text{speed}}
\end{equation}
The cost-to-go term $J_{\text{go}}$ is written as in \cite{yu2025spatial} by using a third order polynomial to fit the remaining race track length. 
An additional cost term $J_{\text{speed}}$ is introduced to control the desired speed. This term is expanded as: 
\begin{equation}
    J_{\text{speed}}  = \int^{t_f}_{t_0} w_{u_x}\left(u_x - \sqrt{\frac{(\mu_f + \mu_r) g}{2\kappa}K_{\kappa}}\right)^2 dt
\end{equation}
where $w_{u_x}$ is the weight, $u_x$ is the longitudinal speed of the vehicle, $\mu_f$ and $\mu_r$ are the friction coefficients on the front and rear tires respectively, $g$ represents the gravitational acceleration, $\kappa$ represents the largest curvature of the centerline within the 200 m track length relative to the current vehicle position, and $K_\kappa$ is a parameter accounting for the vehicle's deviation from the centerline during cornering maneuvers. 
By adding $J_{\text{speed}}$, the algorithm can decelerate sooner when it is approaching a corner.

\subsubsection{Solving Implementation}  This study employs a fixed horizon of 4 s, with 25 points uniformly distributed along the trajectory. 
Between each time step, the backward Euler integration scheme is used for its computational efficiency. 
The nonlinear optimization problem is transcribed into a nonlinear program using NLOptControl \cite{nloptcontrol} and solved with IPOPT \cite{wachterIpopt}. 
This implementation allows the algorithm to plan the trajectories at a frequency of 10 Hz.

\subsection{Haptic Shared Control} \label{sec:HSC}
The haptic shared control framework, depicted in Fig.~\ref{fig:framework}-D, utilizes haptics on the steering wheel as a teaching method. 
It facilitates smooth communication between autonomy and the human driver, enabling seamless transfer of control authority.

Depicted in Fig.~\ref{fig:framework}-D3, the torque generator is a component of the haptic shared control system. 
It receives as inputs the road alignment torque $\tau_{c}$ from the plant model (Fig.~\ref{fig:framework}-B), the steering angle $\delta_{w}$ from the steering wheel (Fig.~\ref{fig:framework}-D1), the steering angle reference $\delta_{w,\text{ref}}$ from the autonomy (Fig.~\ref{fig:framework}-A), and the human torque $\tau_h$ (Fig.~\ref{fig:framework}-D2). Subsequently, it produces the shared control torque $\tau_{sc}$.
Human subjects perceive the shared control torque $\tau_{sc}$ on the steering wheel and adjust their torque output $\tau_h$.
The resultant torque on the steering wheel $\tau_s$, comprising both the shared control torque $\tau_{sc}$ and the human torque $\tau_{h}$, ultimately determines the steering command $\delta_w$, thereby forming a closed loop.

The torque generator is responsible for generating haptic feedback for human subjects. 
In this design, three types of torques are considered:
\begin{enumerate}
    \item $\tau_{\text{ch}}$ (Counter-human torque): This torque is a counter torque for human input $\tau_h$; i.e., ideally $\tau_{\text{ch}} = -\tau_h$. When fully utilized, $tau_{\text{ch}}$  removes the human's influence on the wheel
    \item $\tau_{\text{c}}$ (Road alignment torque): This torque mimics the natural road feel.
    \item $\tau_{\text{PID}}$ (Autonomy torque): Given the current steering angle $\delta_w$ and the reference $\delta_{w,{\text{ref}}}$ from autonomy, a PID controller acts on their difference and outputs the corresponding torque to track the desired steering angle. When $\tau_{\text{PID}}$ is the only torque in the torque generator, the tracker can accurately trace the trajectory designed by the MPC, resulting in minimal time-to-goal performance.
\end{enumerate}

The shared control torque output is then expressed as:
\begin{equation}
    \label{eq:tau_shared_control}
    \tau_{sc} = \alpha  \tau_{\text{PID}} + \beta  \tau_{\text{c}} + \gamma  \tau_{\text{ch}}
\end{equation}
Here, $\alpha$, $\beta$ and $\gamma$ are parameters that control the different phases of assistance. 
Their values vary as a function of the assistance level as depicted in Fig.~\ref{fig:framework}-F.
This design aims to first guide the driver using autonomy and then gradually phase it out, reintroducing intrinsic driving feel to encourage skill development and independent performance. 
Specifically, it has the following properties:
\begin{enumerate}
    \item At assistance level 100, the steering wheel torque $\tau_\text{s}$ equals to $\tau_{\text{PID}}$:
    \begin{equation}
        \tau_\text{s} = \tau_{\text{sc}} + \tau_{\text{h}} = \tau_{\text{PID}} + \tau_{\text{ch}} + \tau_{\text{h}} = \tau_{\text{PID}}
    \end{equation}
    because, ideally, the counter-human torque $\tau_\text{ch}$ should counteract the human input torque $\tau_h$, giving autonomy full control authority.
    \item Between assistance levels from 100 to 60, the counter-human torque gradually diminishes. 
    The algorithm retains its ability to autonomously complete the track with optimal performance, if the driver minimizes their disturbance to the steering wheel. 
    At this stage, the assistance level is designated as \textbf{High}.
    \item At assistance levels ranging from 60 to 35, the counter-human torque $\tau_{\text{ch}}$ is zero, and the autonomy torque $\tau_{\text{PID}}$ begins to decrease.
    At assistance level 35, the autonomy can still complete the track on its own, albeit with diminished performance.
    At this stage, the assistance level is classified as \textbf{Medium}.
    \item From assistance level 35 to 0, the autonomy torque struggles to guide the vehicle effectively to the endpoint and gradually diminishes. 
    Meanwhile, the road alignment torque increases to replicate the realistic torque feedback experienced in racing. 
    At this stage, the assistance level is labeled as \textbf{Low}.
    \item Finally, at assistance level 0, the autonomy torque vanishes entirely, leaving only the road alignment torque perceptible to the human drivers. At this stage, drivers gain full control of the steering task. 
\end{enumerate}

\subsection{Performance Evaluation Metrics}
To evaluate the racing performance of a human driver quantitatively, a scoring mechanism is developed, taking into account three main aspects of racing: lap completion percentage $p_l$, lap time $T_l$, and boundary violation area $A_{B}$.

The simulation terminates if:
\begin{enumerate*}
    \item the vehicle's absolute yaw rate is higher than 1.2 rad/s or the absolute lateral velocity is larger than 8 m/s, vehicle is spinning or sliding excessively; or 
    \item the vehicle deviates more than 15~m from the track boundary.
\end{enumerate*}

If the driver reaches the checkered flag, the lap completion percentage is $p_l=1$. 
Conversely, if any of the termination criteria is triggered, the driver's ending position is projected back onto the track, identified as the point of failure, and utilized to evaluate the lap completion percentage.
For example, if the driver fails in the middle of the track, the lap completion percentage equals 50\%.

The lap time $T_l$ is identical to the finishing time when the track is completed ($p_l = 1$). 
To differentiate between fast drivers and early failing drivers, the lap time is adjusted using the projection shown below:

\begin{enumerate}
    \item Get the current simulation time $T_{l}^\prime$
    \item Find the nearest point on the track $P_f$.
    \item Calculate the time $T_{\text{mpc}}^\prime$ that the autonomy needs to reach $P_f$.
    \item Calculate the time $T_{\text{mpc}}$ that the autonomy needs to finish the whole lap.
    \item Calculate the projected lap time as $T_l =  \frac{T_{\text{mpc}}}{T_{\text{mpc}}^\prime}   T_{l}^\prime$
    % \begin{equation}
    %     T_l =  \frac{T_{\text{mpc}}}{T_{\text{mpc}}^\prime}   T_{l}^\prime
    % \end{equation}
\end{enumerate}

The boundary violation area $A_{B}$ is the total area between the track boundary and the path of the vehicle's center of gravity (CG) when the CG is outside of the boundaries.
The metric $A_{B}$ is the actual total boundary violation area if the driver finishes the lap. 
If the driver fails to complete the lap, the projection of this metric to the entire lap follows the steps below:
% This metric is also projected to the entire map following a slight difference compared to lap time since the expert solution does not exceed the boundary.
\begin{enumerate}
    \item Get the current boundary violation area $A_{B}^\prime$
    \item Find the lap completion percentage $p_l$.
    \item Calculate the projected boundary violation area as $A_{B}~=~\frac{A_{B}^\prime}{p_l}$.
    % \begin{equation}
    %     A_{B} =  \frac{A_{B}^\prime}{p_l}
    % \end{equation}
\end{enumerate}

% The rationale for projecting the metrics to the entire track stems from the metric scores being subsequently determined based on the metric value for the entire track. 
% Denoting the two metrics \Tulga{which two metrics?} using a placeholder $\bullet$, the best-case value $\bullet_{\text{best}}$ (from autonomy solution) and the worst-case value $\bullet_{\text{worst}}$ (from pilot study) are identified. \Tulga{This sentence most likely needs to be revised for clarity. See my attempt below.}
For a given metric $T_l$ or $A_{B}$, denoted with a placeholder $\bullet$, a best and worst case value are determined from the autonomy and pilot studies, respectively. 
In both metrics, the smaller the value is, the better the performance is. 
The score for the given metric is calculated as:
\begin{equation}
    S_{\bullet} = \begin{cases}
        100 \quad \quad \quad \quad \quad \quad \text{if} \,\, \bullet < \bullet_{\text{best}} \\
        0 \quad \quad \quad \quad \quad \quad \quad \text{if} \,\, \bullet > \bullet_{\text{worst}} \\
        100  \frac{\bullet_{\text{worst}} - \bullet}{\bullet_{\text{worst}} - \bullet_{\text{best}}} \quad \; \; \text{otherwise}
    \end{cases}
\end{equation}
Then, the overall racing score is calculated as: 
\begin{equation}
    S_R = p_l  \left(w_t S_{T_l} + w_a S_{A_{B}}\right)
\end{equation}
Here $w_t=0.7$ and $w_a=0.3$ are the weights for the score on lap time and boundary violation area, respectively. 
These weights are derived from prior pilot studies and are distributed to prioritize lap time over boundary violations in the rating system.
The racing score is thus bounded between 0 and 100 and the higher the score is, the better the performance is.

\subsection{Fading Scheme}
\label{sec:FS}
In the fading scheme, the assistance level for subsequent trials is dynamically adjusted based on the racing score obtained in the current trial.
Three curves are incorporated into the fading scheme, each a function of the trial number.	
The nominal curve, depicted as the blue solid line in Fig.~\ref{fig:framework}-E, is employed to ensure a smooth transition of the assistance level.	
The upper bound, illustrated as the red dashed line in Fig.~\ref{fig:framework}-E, is implemented to ensure that participants encounter sufficiently low assistance levels throughout the entire training session, regardless of their racing scores.
The lower bound shown as the orange dashed line in Fig.~\ref{fig:framework}-E, serves to prevent sudden drops in the assistance level, particularly when participants achieve high racing scores while operating at a high assistance level. 
To facilitate effective learning, the fading scheme is specifically designed to follow an S-curve, as observed in learning curves discussed in \cite{anzanello2011learning}. Equations \eqref{eq:dL} to \eqref{eq:lower_bound} generally adhere to this design while providing sufficient flexibility to tailor learning curves for different subjects.

The process of fading is as follows: 
\begin{enumerate}
    \item In the first trial ($K = 1$), the assistance level $L(K)$ is initialized at 100.
    \item The change of assistance level $dL$ is calculated as:
    \begin{align}\label{eq:dL}
    dL &= \min(\max(\alpha  (S_p - S_K), \, \, -15), \, 25)  \\
    \alpha &=  \begin{cases}
    1.5  \quad \text{if} \,\, (S_p - S_K \leq 0) \\ 0.8 \quad \text{otherwise}
    \end{cases}
    \end{align}
where $S_K$ is the racing score in $K^{th}$ trial, and $S_P$ is the passing score defined as 90. Any score higher than 90 is considered a pass, meaning that the driver at least finished the whole lap with a small boundary violation area, thereby demonstrating a basic idea of racing.
    \item The unregulated next assistance level $L^\prime(K+1)$ is calculated as
    \begin{align}
    L^\prime(K+1) &= w_1 \left(L(K) + dL\right) +  w_2  \mathbb{G}_n(K)
    \end{align}
where $\mathbb{G}_n$ is the nominal curve that is designed to regulate the change of assistance level, and $w_1$ and $w_2$ are the weights. 
    \item The next assistance level is regulated as:
    \begin{equation}
        L(K+1) = \min(\max(L^\prime(K+1), \mathbb{G}_l(K)), \mathbb{G}_u(K))
    \end{equation}
where $\mathbb{G}_l(K)$ and $\mathbb{G}_u(K)$ are the lower and upper bounds on the $K^{th}$ trial, respectively.
    \item If the trial number is larger than 23, then  the assistance level is enforced to be 0.
    \item The process is  repeated from step \#2 to \#5.
\end{enumerate}
The curves $\mathbb{G}_n$, $\mathbb{G}_u$, and $\mathbb{G}_l$  are formulated heuristically based on pilot studies to prevent abrupt changes in assistance level and are defined as:
\begin{align}
    &\mathbb{G}_n(K)  = 50\left(\tanh{(-0.165 (K - 12))} + 1\right)\\
    &\mathbb{G}_u(K)  = 50\left(\tanh{(-0.175 (K - 14.5))} + 1\right)\\
    &\mathbb{G}_l(K) \, =\max\left( 40\left(\tanh{(-0.17 (K - 12))} + 1\right) -5, 0\right) \label{eq:lower_bound}
\end{align}

\section{Experimental Setup}\label{sec:exp_setup}
\begin{figure*}
\centering
\includegraphics[width=0.9\textwidth]{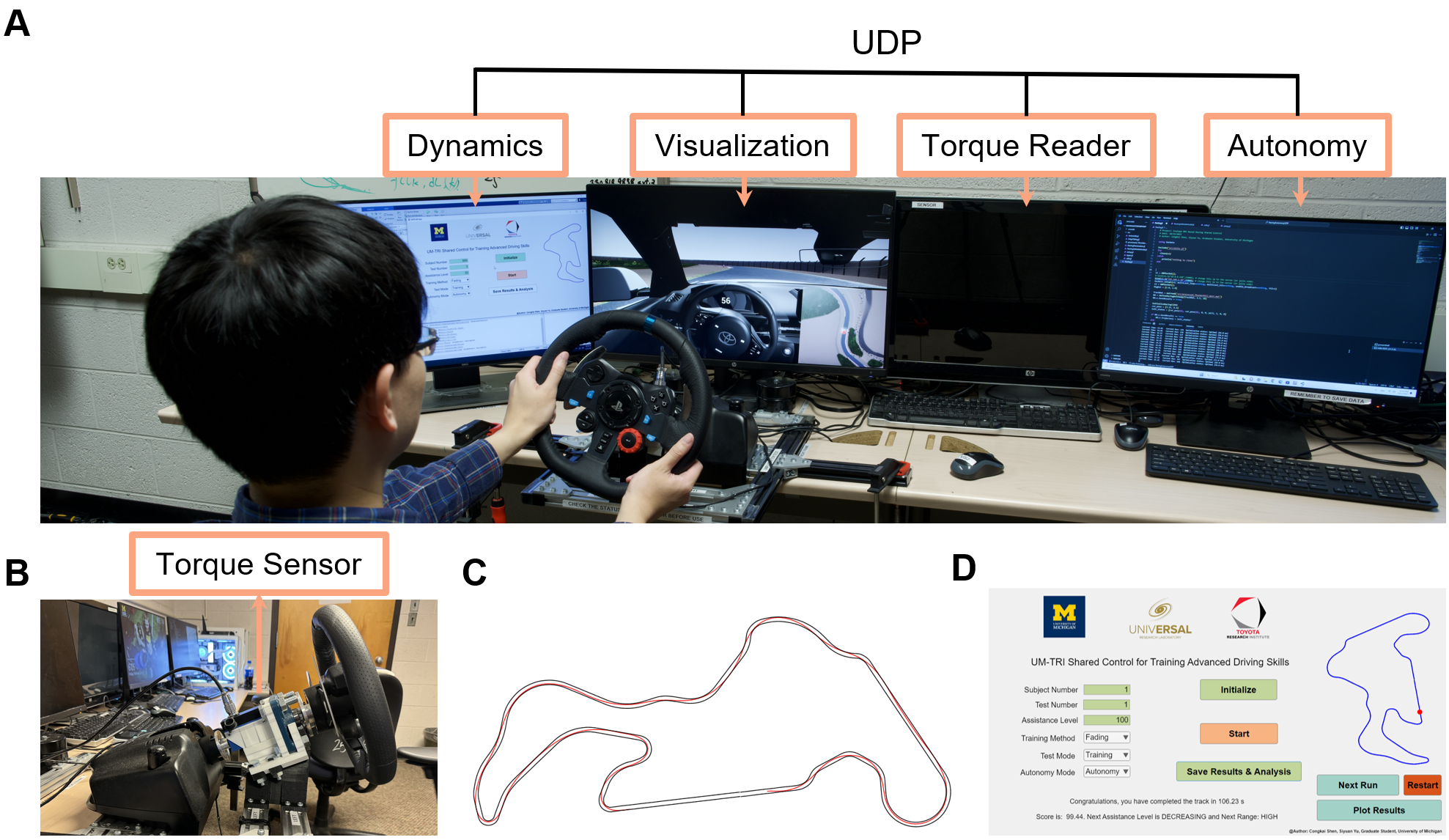}
\caption{Experimental Setup: (\textbf{A}) The simulation with a human driver in the loop. (\textbf{B}) The torque sensor added to the steering column. (\textbf{C}) The track map of Thunderhill West Raceway and the autonomy path without human intervention. (\textbf{D}) The user interface to control the workflow of the testbed and display necessary information to the human driver between trials.}
\label{fig:experimental_setup}
\end{figure*}

\subsection{Human Participants} \label{sec:human_subject}
A total of 48 participants were recruited from the University of Michigan, Ann Arbor. 
Of all the participants, nine identified themselves as female, thirty-eight as male, and one chose not to disclose their gender. 
The subjects were on average 23.3 years old, with a standard deviation of 2.65 years old. 
Thirty-seven participants reported prior experience driving a race car, including Go Karts, and twenty-nine indicated playing racing games at least once a year. 
Thirty-two of them have participated in shared control related tests before. 
This study adhered to the American Psychological Association's ethical guidelines and received approval from the Institutional Review Board at the University of Michigan (Application No. HUM\#00246155).

\subsection{Experimental Apparatus and Stimuli}
As shown in Fig. \ref{fig:experimental_setup}, the experiment employs a fixed-base driving simulator for controlling a simulated race car. 
The human operates a Logitech steering wheel to navigate the vehicle along the racetrack. 
For the measurement of human torque as an input for haptic shared control, a torque sensor is installed on the steering column, depicted in Fig. \ref{fig:experimental_setup}-B. 
The testbed employs four computing systems for vehicle dynamics, visualization, torque measurement, and autonomy. 
These systems communicate via the User Datagram Protocol (UDP) over the ethernet.	

The track used in the autonomy and visualization represents Thunderhill West Raceway without elevation changes.  
A first-person view from the driver's seat in the cockpit is visualized in the virtual environment. 
The current speed in miles per hour and a small map from the aerial view are also shown on the screen. 
The visualization is created in Unreal Engine 4 and is capable of showing the vehicle poses in real-time with a frame rate of approximately 60 fps.
Given that human drivers cannot perceive physical accelerations in fixed-base simulators, additional illustrations of small hills and grass fields are incorporated into the virtual environment to assist with the sensation of speed for the driver. 
Those features do not intervene with the dynamics and are solely for visualization. 
Fig. \ref{fig:experimental_setup}-C shows the track boundaries (in black) and the autonomy path (in blue), navigated without human intervention.

\subsection{Experimental Procedure}\label{sec:exp_procedure}
The human subject study was conducted from January 2024 to February 2024 at the Walter E. Lay Automotive Laboratory.
Upon arrival at the testing location, participants were initially assigned a random number from one to three, determining their group assignment. 
Next, participants watched a video that explained the entire test procedure and provided information about the performance score; they were only told there would be a score without knowing how the score was calculated. 
They were then instructed to carefully read and sign the consent form, as well as complete an anonymous demographic survey if they chose to participate in the study.	

Then, the participants underwent eight runs of a pre-training test to assess their basic skills in this specific task. 
For all runs in the pre-training test and later in the post-training test, the assistance level described in Sec.~\ref{sec:HSC} was kept at zero. 
Following the completion of each run, the user interface depicted the corresponding racing score and lap time, as illustrated in Fig. \ref{fig:experimental_setup}-D.

Subsequently, the subjects underwent twenty-five training runs to improve their skills in this particular task.
Based on the group number, each subject was randomly assigned to a training group upon arrival.
Regardless of their group, the subjects were instructed to keep their hands on the steering wheel throughout the whole training.
The training protocols specific to each group are as follows:
\begin{enumerate}
    \item \textbf{Group 1}: All training sessions were conducted with zero assistance. 
    Participants did not feel any steering assistance from the autonomy, but only the self-aligning torque from the tires. 
    Thus, this group of participants engaged in a self-learning training phase.
    \item \textbf{Group 2}: All training sessions were conducted at the maximum assistance level of 100. 
    Participants experienced robust steering assistance from the autonomous system throughout the entire training period.
    \item \textbf{Group 3}: The assistance level varied based on the fading scheme introduced in Sec. \ref{sec:FS}. 
    The steering assistance  gradually faded away during the training process. 
    For the \textbf{High} stage, human subjects were advised to exercise caution when intervening with the system. 
    For the \textbf{Medium} stage, human subjects were advised to adapt their interventions in the system while also accounting for the reduced performance due to the decreased torque assistance. 
    For the \textbf{Low} stage, human subjects were advised to assume the dominant role in the shared control interaction, and learn to counteract the road alignment torque while still attaining satisfactory performance.
\end{enumerate}

Between each iteration during the training phase, the racing score and lap time were shown to the participants in all groups. Additionally, for Group 3, the next assistance level phase was also displayed.	

Following the conclusion of the training phase, a ten-minute break was enforced to facilitate the subsequent short-term retention assessment.
Finally, the second testing was conducted, comprising another eight runs without any assistance from autonomy to evaluate the participants' post-training skills.

\section{Experimental Results}\label{sec:exp_res}

The test is separated into three sections. 
The first section is the pre-training session, which evaluates the subject's original racing skill. 
The second section is the training session, of which the results are expounded in Sec.~\ref{sec:train}. 
The last section is the post-training session and the detailed results are shown in Sec.~\ref{sec:post_train}.

Group comparisons are performed utilizing the Brown-Forsythe and Welch Analysis of Variance (ANOVA) test, followed by Dunnett's T3 multiple comparisons as the post-hoc test. 
Significance levels of the results are indicated by asterisks to denote varying degrees of statistical significance. 
Namely, $*$ indicates significance level $p<0.05$, $**$ indicates $p<0.01$ and ns stands for no statistical significance.

\subsection{Pre-training Results} \label{sec:pre_train}
\begin{figure}
\centering
\includegraphics[width=0.3\textwidth]{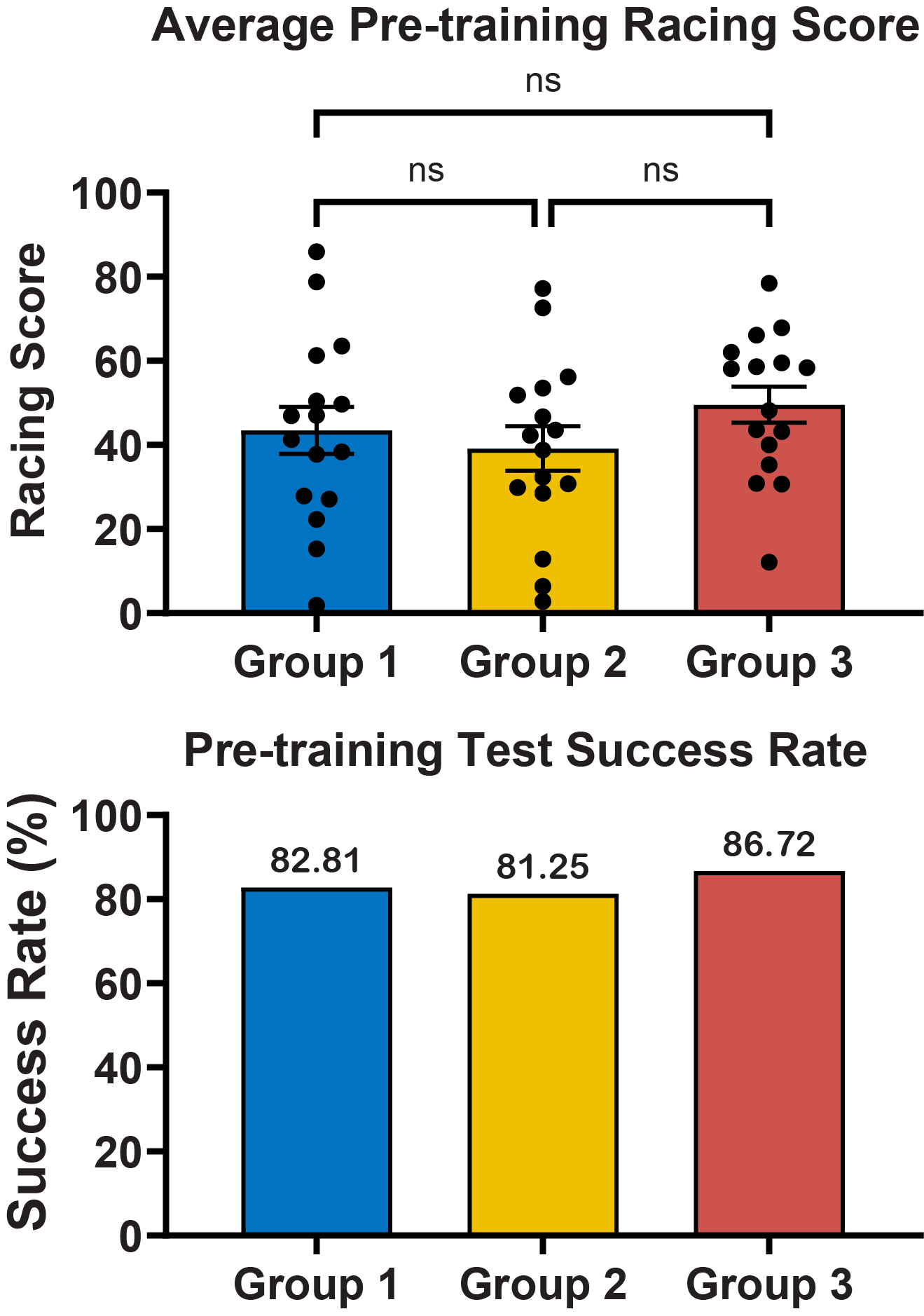}
\caption{Racing scores and success rates of pre-training tests. The black dots represent the real data from individual subjects. The height of each bar indicates the mean value, and the error bars depict the standard error. The brackets above the bar chart indicate the level of statistical significance.}
\label{fig:pretraining_score}
\end{figure}
The pre-training racing scores for the three groups are shown in Fig.~\ref{fig:pretraining_score}. 
Group 1 achieves an average score of 43.48, with a standard deviation (SD) of 22.3. 
Group 2 attains an average score of 39.15, with a SD of 21.16. 
Group 3 obtains an average score of 49.57, with a SD of 17.15. 
There is no statistically significant difference between any pair of groups, suggesting that the initial racing skill levels of the groups exhibit no statistical difference.

The success rate of each group is reported also in Fig.~\ref{fig:pretraining_score}. 
The success rate is calculated as $\frac{N_{c}}{N_{t}} \times 100 \%$, where $N_c$  represents the number of completed trials by subjects, and $N_t$ denotes the total number of trials (128 runs). 
Group 1 achieves 82.21\% success rate, Group 2 reaches 81.25\%, and Group 3 obtains 86.72\%. 
The success rates of the groups hover around 83\%, again suggesting a similarity in their performance.	

Judging from the pre-training score and success rate comparisons, the distribution of initial skill levels among the three groups is considered fair. 

\subsection{Training Results} \label{sec:train}
\begin{figure}
\centering
\includegraphics[width=0.38\textwidth]{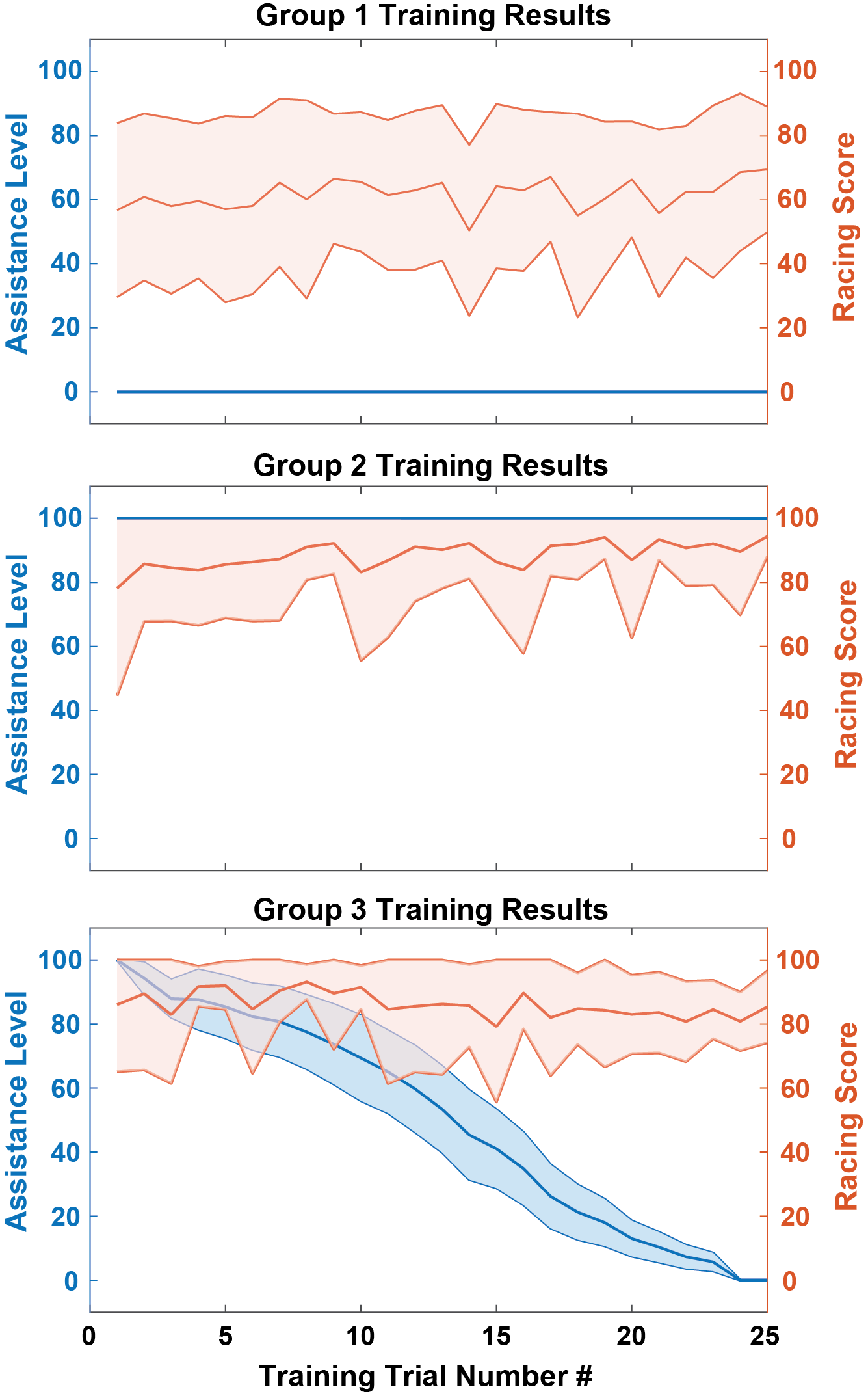}
\caption{Assistance level and racing score throughout the training phase. The center line indicates the mean value of the metric, while the upper and lower lines denote one standard deviation above and below the mean, respectively.}
\label{fig:training_result}
\end{figure}
The training results are shown in Fig.~\ref{fig:training_result}. 
Each subplot illustrates the results obtained by a specific group. 
Each figure displays the progression of the assistance level and racing score over the course of the trials. 

\begin{enumerate}
    \item \textbf{Group 1:} Group 1, designated as the self-learning group, consistently experienced an assistance level of 0 throughout the experiment, depicted by a solid blue line in Fig.~\ref{fig:training_result}. The average racing score of Group 1 starts from around 60 and ends at around 70. The SD remains large along the trial number. The non-convergence of Group 1's racing score indicates that it is still in transient by the end of the training. Thus, this group could benefit from more training runs.

    \item \textbf{Group 2:} Group 2, designated as the full assistance group, consistently experienced an assistance level of 100. Group 2 exhibits an average racing score that begins around 80 and levels around 95 at the end with the aid of full autonomy assistance. 
    % This suggests that, aided by autonomy, participants in Group 2 complete the entire lap, albeit with minimal boundary violations, and demonstrate a fundamental understanding of executing race lines during corners. \Tulga{How can we claim that they demonstrate any fundamental understanding when they are strongly guided by autonomy all the time? I don't believe we can support such a claim at this point.}
    % Note that any score larger than 90 is considered a pass, meaning that the subject at least finished the whole lap with the small boundary violation area and has a basic idea of how to perform race lines on hairpin turns. The SD is not monotonically decreasing but it shows the trend of convergence. \Siyuan{This should be in the fading scheme}. 

    \item \textbf{Group 3:} For Group 3 participants, the assistance level gradually decreases from 100 to 0, and the upper and bottom lines are bounded within a reasonable range due to the regularization of upper bound $\mathbb{G}_u$ and lower bound $\mathbb{G}_l$  as described in Sec.~\ref{sec:FS}. 
    Similar to Group 2, with the 100 assistance level, the racing score starts at around 80. 
    Instead of showing an increase as Group 2, the racing score persists around 85 throughout the duration of the fading assistance level. 
    Between trials 1 and 10, the average assistance level ranges from high (100) to medium (60), with a noticeable convergence trend observed in the SD of the racing score. 
    Subsequent to the 10th trial, the average assistance level transitions into the medium range, while the SD of the racing score diverges back to the initial level. 
    With the increase in trial number, the SD begins to converge again. 
    This turning point underscores the significance of the fading scheme, demonstrating that despite achieving commendable performance at a high assistance level, maintaining such performance under medium and low assistance levels becomes challenging.
\end{enumerate}

\subsection{Post-training Results} \label{sec:post_train}
\begin{figure*}
\centering
\includegraphics[width=0.9\textwidth]{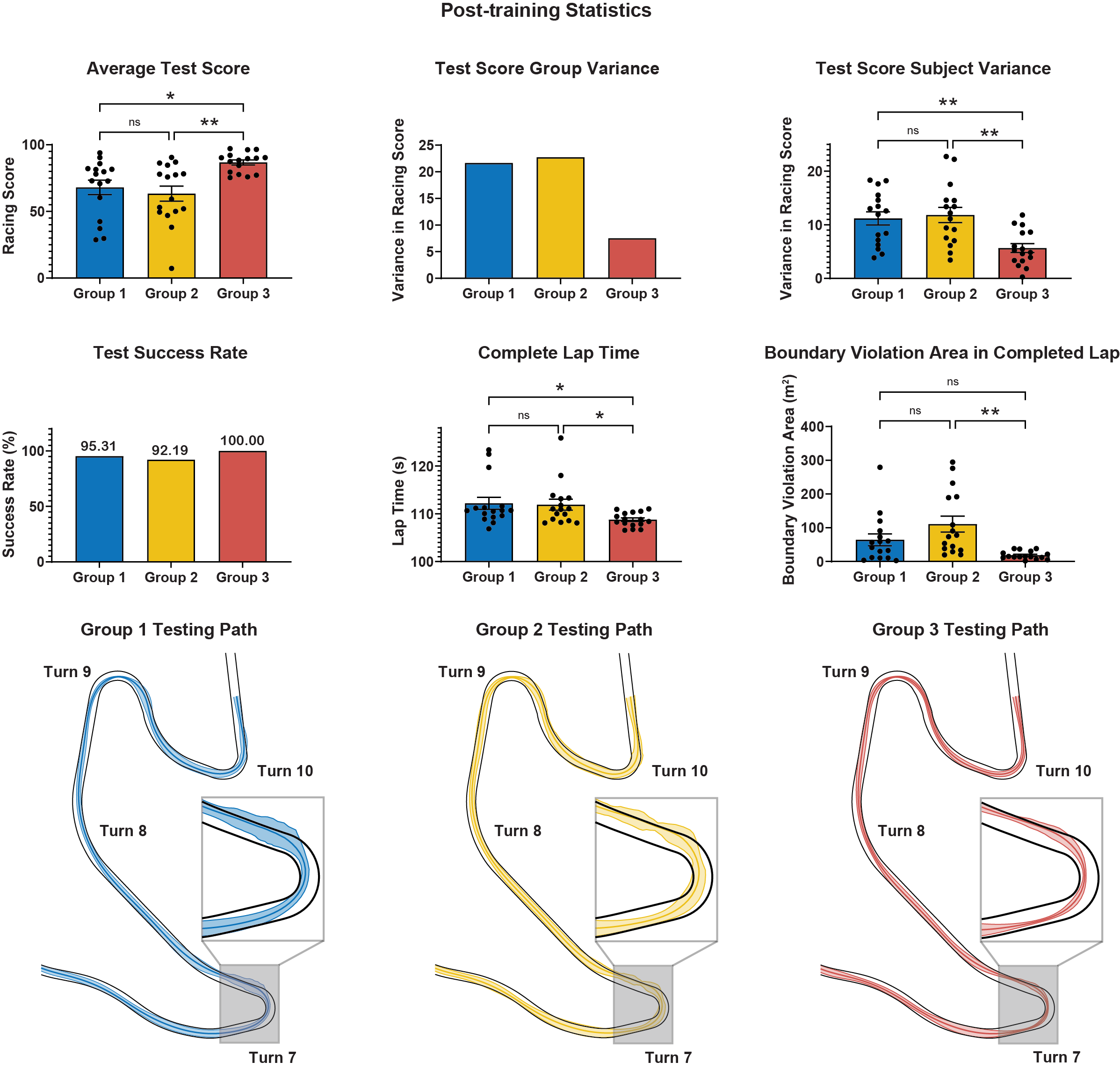}
\caption{Post-training results. (\textbf{A}) Statistical analysis of metrics in the post-training test. (\textbf{B}) Example vehicle paths. Middle line is the mean path on the track and the two lines bounding the middle line denote one standard deviation from the mean path. The grey area magnified the driver performance in turn 7.}
\label{fig:testing_result}
\end{figure*}

The ANOVA tests for the post-training results and the path distribution plots are shown in Fig.~\ref{fig:testing_result}. 
The summary of the examination of six metrics is provided below.

\begin{enumerate}
    \item \textbf{Average Racing Score:}  Compared to Fig.~\ref{fig:pretraining_score}, all groups demonstrate improvements in their average racing scores after the training sessions. 
    In Fig.~\ref{fig:testing_result}, Group 2 exhibits the lowest average score of 63.38. 
    Group 1's average score is slightly higher than that of Group 2 by 4.65. 
    Group 3 shows the highest average score of 86.77. From the posthoc test, Group 1 and 2 show no statistical difference, while Group 3 has statistical differences with both Group 1 and 2. 
    In the plot, each subject's average racing score is depicted with a black dot.
    Thus, the results imply that after the training, Group 3 subjects reach the highest average racing score.

    \item \textbf{Score SD within Group:}  The SD of the average racing scores within each group is denoted as the score SD within group. 
    Group 2 exhibits the highest score SD within group at 22.74, whereas Group 1 has a smaller score SD within group than Group 2 does by 1.08. 
    Group 3 has the smallest score SD within group of 7.5. 
    Thus, Group 3 demonstrates the highest consistency in performance across all subjects within the group.
    
    \item \textbf{Average of Score SD within Subjects:}  This metric represents the average variability within individual subjects within the group.	
    More specifically, the SD for each subject is computed based on their 8 post-training racing scores, and the average score SD within subjects for a given group is calculated as the mean of the SDs of all the subjects in the group. 
    Group 3 exhibits the lowest average of score SD within subjects and demonstrates statistically significant differences from both Groups 1 and 2, whereas Groups 1 and 2 do not differ significantly from each other. 
    Therefore, it is inferred that the racing scores within Group 3 subjects are more consistent. 
    In the plot, the black dots represent the score SD of each subject.

    \item \textbf{Success Rate:} Compared to Fig.~\ref{fig:pretraining_score}, following the training session, the success rates of Groups 1 and 2 have surpassed 90\%, yet instances of failure persist. 
    Meanwhile, all subjects in Group 3 are capable of completing every lap in the testing and reaching a 100\% success rate. 
    This demonstrates that the subjects in Group 3 show better capability to handle the vehicle safely on the track without losing control.

    \item \textbf{Average Lap Time in Completed Laps:} The metric for each group is computed as the average of its subjects' average lap times. 
    Any failures experienced by subjects are excluded from the calculation of their average lap time. 
    Group 3 has the fastest completed lap time of 108.76 s with an SD of 1.51 s, showing statistical differences with Groups 1 and 2. 
    Group 1 exhibits the slowest average lap time, clocking in at 112.19 s, with an SD of 5.01 s. 
    Group 2 shows an average lap time of 111.92 s, accompanied by an SD of 4.57 s. 
    There exists no statistically significant difference between Groups 1 and 2. 
    Thus, subjects in Group 3 demonstrate superior lap time performance compared to the other two groups, suggesting an ability to better control the vehicle at the limits of handling.
    In the plot, each subject's average lap time is denoted with a black dot.

    \item \textbf{Average Boundary Violation Area in Completed Laps:} Group 2 has the largest boundary violation area of 110.78~$\text{m}^2$ with an SD of 94.33 $\text{m}^2$. 
    Compared to Group 2, Group 1 demonstrates a smaller boundary violation area, showing a difference of 46.44 $\text{m}^2$, and an SD of 70.82 $\text{m}^2$. 
    Group 3 exhibits the smallest boundary violation area, measuring at 18.52 $\text{m}^2$, accompanied by an SD of 11.83 $\text{m}^2$. 
    A statistically significant difference exists only between Groups 3 and 2. 
    A marginal difference between Groups 3 and 1 is observed, indicated by a p-value of 0.0621. 
    Unlike other metrics, the difference between Groups 1 and 2 is noticeable in the boundary violation area, albeit not statistically significant. 
    This phenomenon can be attributed to the unfamiliarity of subjects in Group 2 with the abrupt transition from assistance level 100 to 0. 
    Consequently, upon the disappearance of haptic guidance and the introduction of centering torque, subjects experience confusion and hesitation, leading to delayed turns and substantial deviations from the boundary.

\end{enumerate}

In Fig.~\ref{fig:testing_result}, the path distributions of all three groups are demonstrated. 
The turns shown are the most challenging portions of the Thunderhill West Raceway, where turns 7, 9, and 10 are hairpin bends, and turn 8 is a high-speed corner. 
The distribution is composed of three lines. 
The middle line represents the mean path, and the other two lines form the distribution boundary around the mean path within one standard deviation. 
On turns 7, 9, and 10, the path distribution of Group 3 converges to the outer lane before entering the apex, resulting in a small boundary violation after exiting the apex. 
On the contrary, Groups 1 and 2 show large variances before entering the apex. 
The large variance in the turn-in point causes a large variance in the exit point, leading to a large boundary violation area. 
The comparisons in path distributions partially explain why Group 3 has the smallest boundary violation, the fastest lap time, and thus the highest racing score.

Based on the aforementioned findings, three important conclusions are drawn:
\begin{enumerate}
       % \item Haptic shared control alone is not sufficient for good training
    \item Haptic shared control alone is insufficient for effective training, as demonstrated by the lower post-training skill levels observed in the full assistance and self-learning groups.
        
    \item With full, always-on haptic assistance, no additional benefits are provided compared to self-learning, as reflected in the comparable post-training skill levels of both groups.
    
    \item Haptic assistance with a fading scheme in shared control shows promise for fostering long-term benefits in high-performance driving scenarios, evidenced by the superior short-term retention of advanced driving skills in Group 3 compared to the two benchmark groups.
    % \item \Tulga{We may be able to enrich our conclusions. After the introduction is re-written, we can come back to this list and see if the conclusions are addressing the re\textbf{}search questions. For instance, it may be good to say explicitly that haptic shared control alone is not sufficient for good training, as results with full, always-on haptic assistance do not lead to benefits over self training. How the haptic assistance is applied also needs to be carefully designed as demonstrated with the fading scheme. This way we can bring the level of the terminology back to where we started. Even though this message is not saying anything new, it is making some connections explicit that are otherwise hidden behind the terms full assistance and fading scheme.}\Siyuan{Discussion here}
\end{enumerate}

\section{Conclusion}\label{sec:conclusion}
This work develops a novel haptic shared control framework and utilizes it for teaching advanced driving skills. 
The novelty lies in tailoring a performant autonomy formulation for shared control, and creating a haptic shared control framework along with a fading scheme to enable efficient training. 
A human subject study, coupled with statistical analysis, indicates that with the new framework, drivers attain enhanced racing skills, encompassing improved racing performance (both in lap time and boundary violation) and greater consistency in individual performance.	
Additionally, drivers trained by the fading scheme reach the same skill level more consistently than the benchmark training frameworks.
% \Tulga{What is the difference between consistency in 'greater consistency' and consistency in 'more consistent skill levels'? Is the last sentence redundant?}\Congkai{Changed the description.
% The first consistency (Average of Score SD within Subjects) refers to the consistent individual performance. The second consistency (Score SD within Group) refers to the group variance, meaning that drivers trained by the fading scheme reach the same skill levels more consistently than the other two groups.}

To the best of the authors' knowledge, this study is the first to utilize haptic shared control as a teaching tool for advanced driving skills.
This framework can be extended beyond applications of racing to provide valuable training in other challenging cases relevant to normal driving scenarios such as imminent collision avoidance.	
% Consequently, this framework represents a significant advancement in validating the notion that haptic feedback is beneficial in teaching.

The findings should be viewed in light of the following limitations. 
Firstly, the design of the fading scheme is heuristic and remains fixed for all subjects.	
For future work, appropriate human modeling in racing would provide invaluable insight into tailoring the design of the fading scheme to each individual. 
Secondly, although the fading scheme changes between training runs, it remains constant within a single run.	
An adaptive haptic shared control system could be employed to dynamically adjust the assistance level online, enhancing its efficacy for teaching purposes.	
Thirdly, the algorithm could be further enhanced by incorporating additional feedback modalities, such as visual and auditory cues, to facilitate more efficient instruction.

\bibliographystyle{IEEEtran}
\bibliography{IEEEfull}

\begin{IEEEbiography}[{\includegraphics[width=1in,height=1.25in,clip,keepaspectratio]{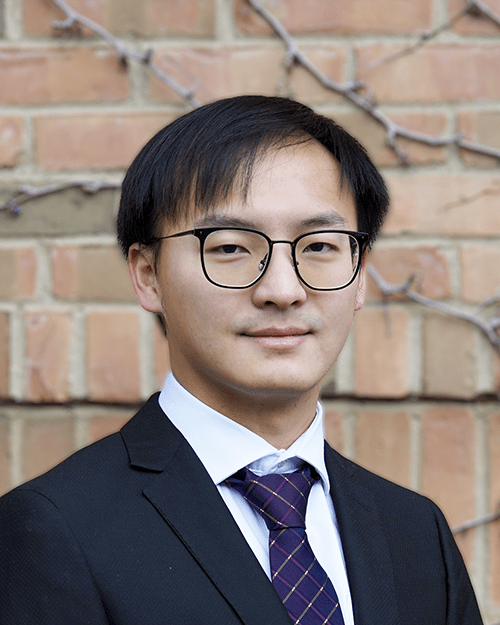}}]{Congkai Shen}
 earned his Bachelor of Science in Electrical and Computer Engineering from Shanghai Jiao Tong University, China, in 2020. Subsequently, he completed his Bachelor of Science and Master of Science in Mechanical Engineering at the University of Michigan, Ann Arbor, in 2020 and 2022, respectively. Currently, he is pursuing the Ph.D. degree in Mechanical Engineering at the University of Michigan, Ann Arbor. His research focuses on modeling, system identification, motion planning, and control, in the context of vehicle systems.
\end{IEEEbiography}

\begin{IEEEbiography}[{\includegraphics[width=1in,height=1.25in,clip,keepaspectratio]{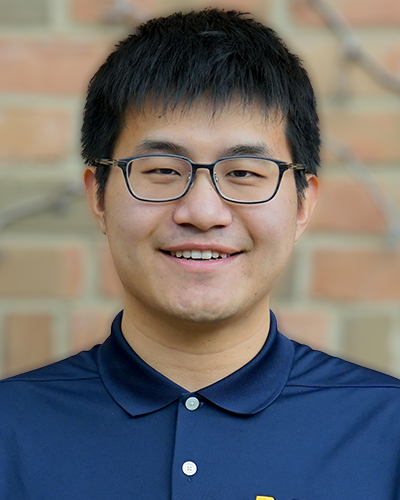}}]{Siyuan Yu}
received his B.S.E in Electrical and Computer Engineering from Shanghai Jiao Tong University, China, in 2020 and the B.S.E and M.S.E in Mechanical Engineering from University of Michigan, Ann Arbor in 2020 and 2022, respectively. He is currently pursuing the Ph.D. degree in Mechanical Engineering at the University of Michigan, Ann Arbor. His research interests include modeling, system identification, motion planning and control, with respect to vehicle systems.
\end{IEEEbiography}

\begin{IEEEbiography}[{\includegraphics[width=1in,height=1.25in,clip,keepaspectratio]{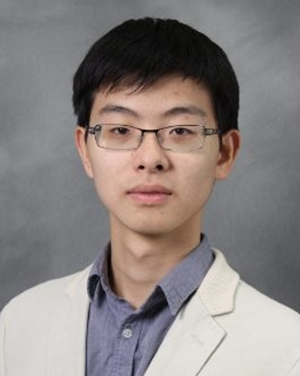}}]{Yifan Weng}
received the B.S. degree from the Shanghai Jiao Tong University, China, in 2016 and the M.S. and Ph.D. degrees from the University of Michigan, Ann Arbor, MI USA, in 2018, and 2022, respectively, all in Mechanical Engineering. His research interests include shared control and motion planning, and control, with applications to vehicle systems. 
\end{IEEEbiography}

\begin{IEEEbiography}[{\includegraphics[width=1in,height=1.25in,clip,keepaspectratio]{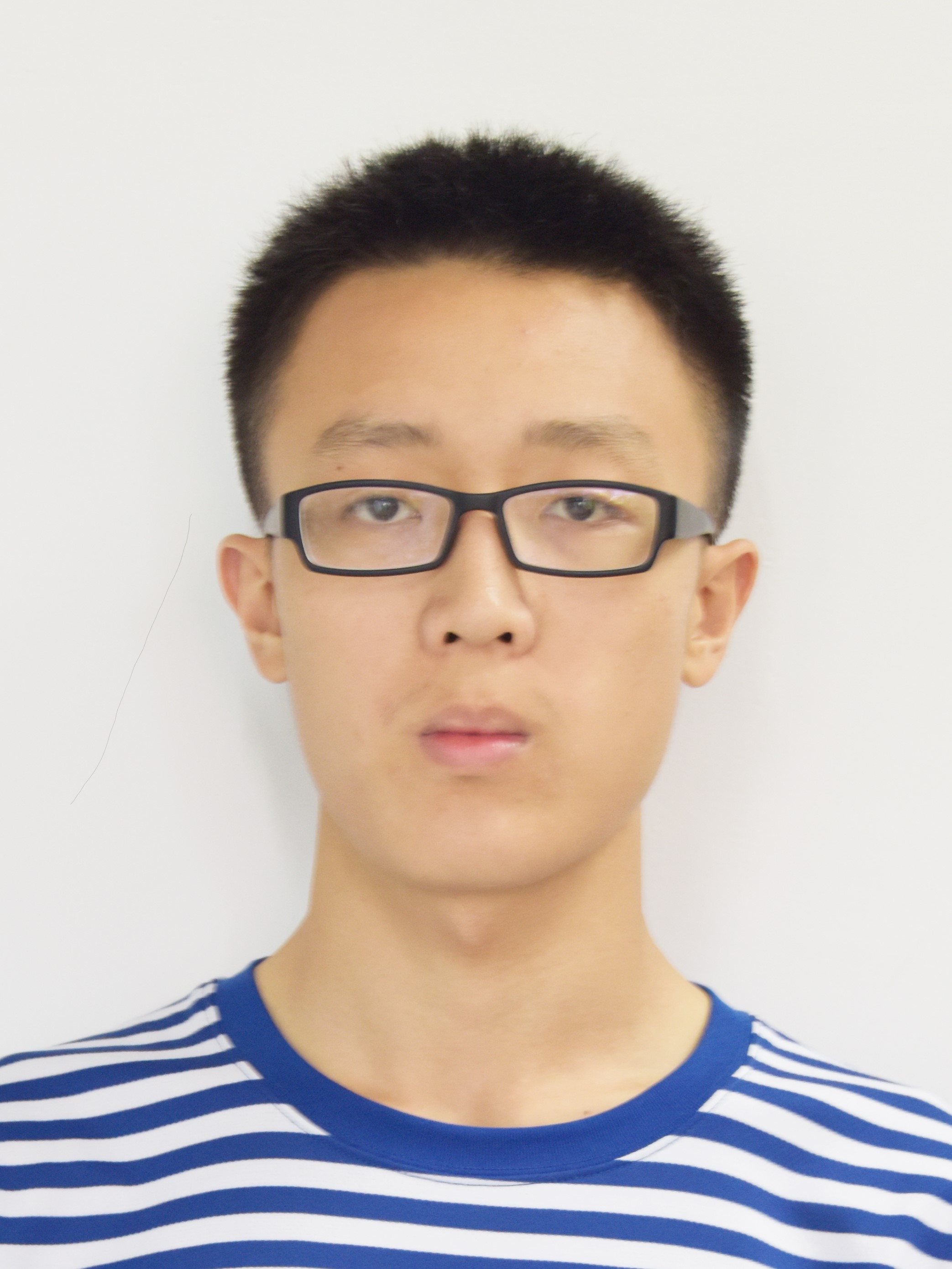}}]{Haoran Ma}
received his B.S.E in Mechanical Engineering from Sichuan University, China and University of Michigan, Ann Arbor in 2024. He is currently pursuing the Ph.D. degree in Mechanical Engineering at the University of Michigan, Ann Arbor. His research interests include human factors, modeling, simulation, and control of vehicle systems.
\end{IEEEbiography}

\begin{IEEEbiography}[{\includegraphics[width=1in,height=1.25in,clip,keepaspectratio]{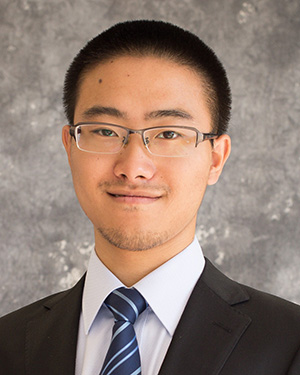}}]{Chen Li}
received the B.S degree in Electrical Engineering from Michigan Technological University, Houghton, MI, USA, in 2015, the M.Eng degree in Electrical Engineering: Systems in 2017, and the Ph.D. degree in Mechanical Engineering in 2023 from the University of Michigan, Ann Arbor, MI, USA, where he is currently a post-doctoral scholar. His research interests include driver modeling, shared control, and human-autonomy interaction.
\end{IEEEbiography}

\begin{IEEEbiography}[{\includegraphics[width=1in,height=1.25in,clip,keepaspectratio]{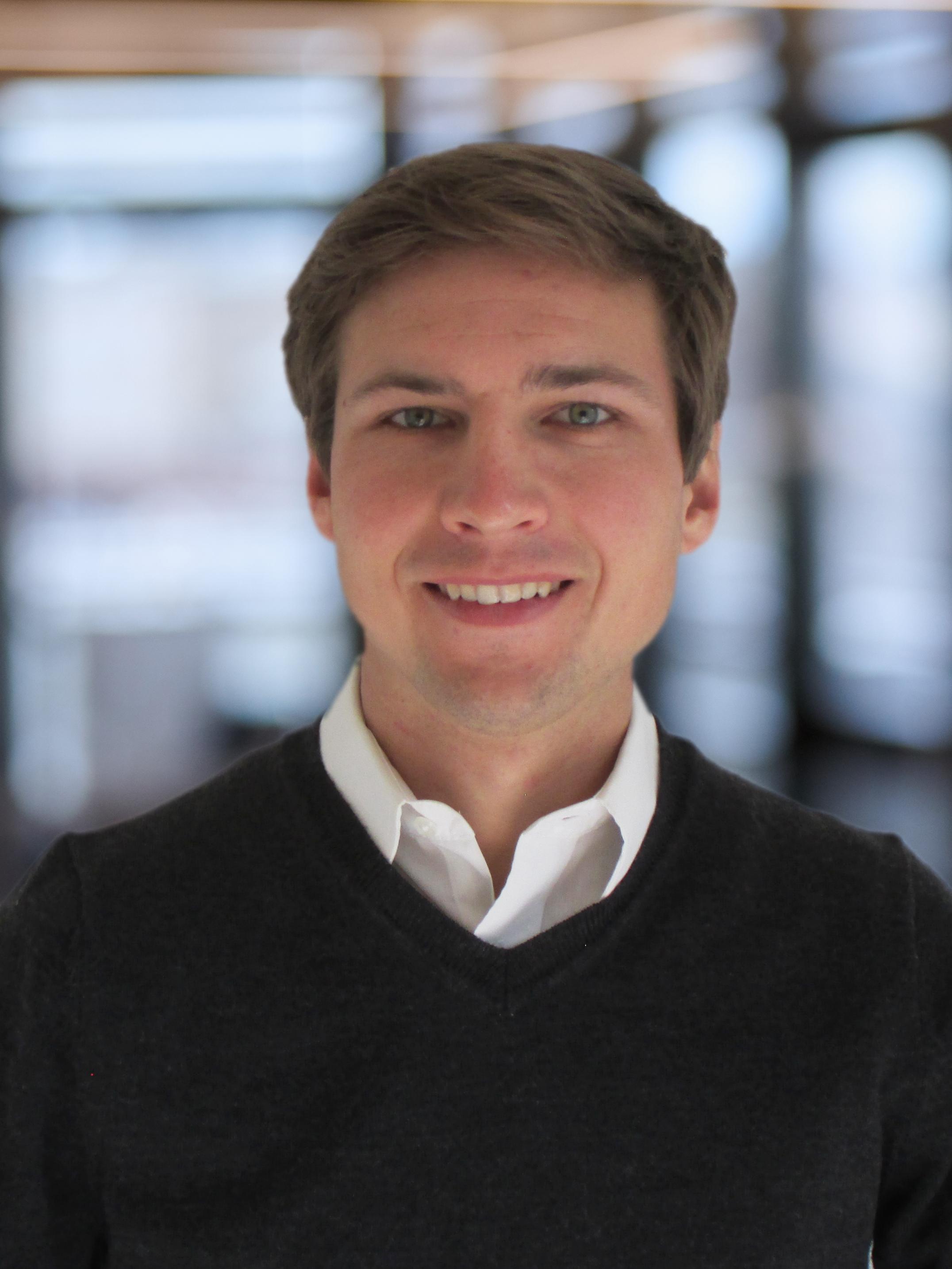}}]{John Subosits}
received his B.S.E. degree in mechanical and aerospace engineering from Princeton University and M.S. and Ph.D. degrees in mechanical engineering from Stanford University.  Currently, he leads the Extreme Performance Intelligent Control group at Toyota Research Institute (TRI).  His research interests include algorithms for vehicle control that match the performance, robustness, and adaptability of the best human (racing) drivers.
\end{IEEEbiography}

\begin{IEEEbiography}[{\includegraphics[width=1in,height=1.25in,clip,keepaspectratio]{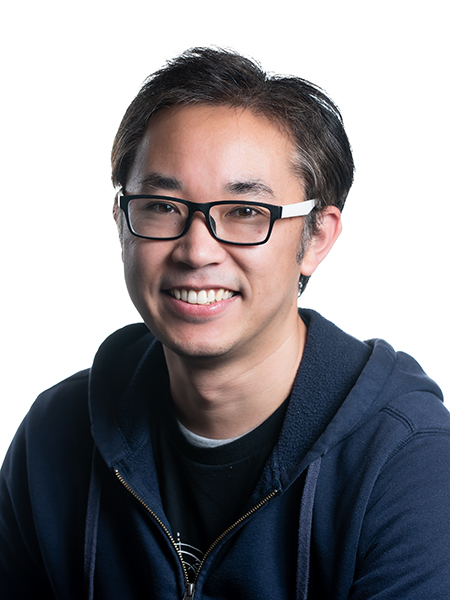}}]{Hiroshi Yasuda}
received a B.E. from the Tokyo University of Science, Japan, in 2003, and M.E. and Ph.D. in engineering from the Tokyo Institute of Technology, Japan, in 2005 and 2008. He is currently a Staff Researcher and HMI tech lead at the Toyota Research Institute in Los Altos, California. His research interests include HMIs for advanced safety systems, and augmented/mixed reality for vehicles.
\end{IEEEbiography}

\begin{IEEEbiography}[{\includegraphics[width=1in,height=1.25in,clip,keepaspectratio]{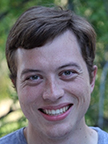}}]{Michael Thompson}
received a B.S. degree in aerospace engineering from the University of Notre Dame and a M.S. degree in aerospace engineering from Stanford University. He is currently a member of the Extreme Performance Intelligent Control group at Toyota Research Institute. His research interests include vehicle dynamics modeling and control algorithms for high performance autonomous vehicles
\end{IEEEbiography}

\begin{IEEEbiography}[{\includegraphics[width=1in,height=1.25in,clip,keepaspectratio]{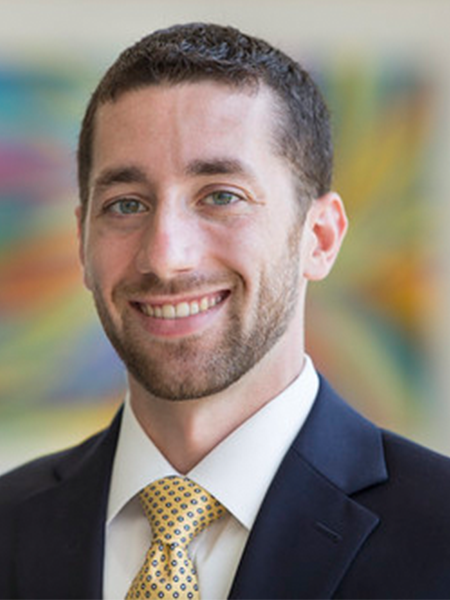}}]{James Dallas}
received the B.S. degree from the Pennsylvania State University, University Park, PA USA, and the M.S. and Ph.D. degrees from the University of Michigan, Ann Arbor, MI USA, in 2017, 2018, and 2021, respectively, all in Mechanical Engineering. He is currently a Research Scientist with the Toyota Research Institute. His research interests include modeling, system identification, system dynamics and control, shared control, and adaptive and robust optimal control, with applications to vehicle systems.
\end{IEEEbiography}

\begin{IEEEbiography}[{\includegraphics[width=1in,height=1.25in,clip,keepaspectratio]{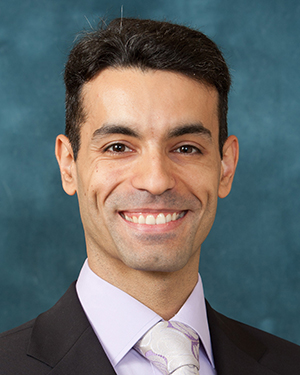}}]{Tulga Ersal}
received the B.S.E. degree from the Istanbul Technical University, Istanbul, Turkey, in 2001, and the M.S. and Ph.D. degrees from the University of Michigan, Ann Arbor, MI USA, in 2003 and 2007, respectively, all in mechanical engineering. He is currently an Associate Professor in the Department of Mechanical Engineering, University of Michigan, Ann Arbor. His research interests include modeling, simulation, and control of dynamic systems, with applications to vehicle and energy systems.
\end{IEEEbiography}

\vfill

\end{document}